\documentclass{article}

\PassOptionsToPackage{numbers}{natbib} % numbered [1] citations, genre standard
\usepackage[preprint]{neurips_2026}

\usepackage[utf8]{inputenc}
\usepackage[T1]{fontenc}
\usepackage{hyperref}
\usepackage{url}
\usepackage{booktabs}
\usepackage{array}
\usepackage{amsmath}
\usepackage{amsfonts}
\usepackage{nicefrac}
\usepackage{microtype}
\usepackage{xcolor}
\usepackage{graphicx}
\usepackage{verbatim}
\usepackage{algorithm}
\usepackage{algpseudocode}

\title{STOCKTAKE: Measuring the Gap Between Perception and Action in LLM Agents with a Fair Oracle}

\author{%
  Sagar Deb\\
  QpiAI\\
  \texttt{sagar.deb@qpiai.tech}\\
  \And
  Ashwanth Krishnan\\
  QpiAI\\
  \texttt{ashwanth.krishnan@qpiai.tech}\\
}

\begin{document}

\maketitle

\begin{abstract}
LLM agents are increasingly evaluated on multi-week decision tasks in
which the state that drives cost is never directly observed. On such tasks
the final cost cannot say why an agent failed. It may have misread the
world, or it may have read the world correctly and still failed to act
(the knowing-doing gap). Existing evaluations cannot separate these two
failures: their reference policies either read privileged information the
agent never sees, or are missing altogether. We introduce STOCKTAKE, a 26-week supply-chain
replenishment benchmark built as a factored partially observable Markov
decision process (POMDP) with six hidden factor processes, designed so that a \emph{fair} reference policy is computable:
an exact Bayes filter per factor drives a rollout policy on the identical
observation stream the agent receives. Scoring each run between a
symptom-blind base-stock floor (0) and this oracle (1) yields a skill
score, and grading each week's written rationale yields a stated-belief
detection lag and a knowing-doing rate, so state estimation and control
are measured separately. On fifty seeds with curated stress profiles,
Claude Sonnet 5, GPT-5.4, DeepSeek-V4-Pro, and Grok 4.5 detect 84--88\%
of hidden failures, typically within a week of onset, yet span skill
scores from $0.62$ to $-0.23$: two of the four end below the
symptom-blind floor while naming factors slightly \emph{faster} than the
two that beat it. The failure has two faces. Where stress persists,
34--43\% of correctly diagnosed stress weeks still end in stockout for
every model (a rate that partly reflects the severity of the weeks models
notice; Section~5). That rate also runs opposite to skill: the two models
under the floor stock out \emph{least} on diagnosed weeks, so
under-response is only one face of the gap, and their traces point to the
other --- responses whose cost exceeds what they protect. The bottleneck
for current agents sits between a correct stated belief and the costly
action it calls for; STOCKTAKE measures both directions of that failure.
\end{abstract}

\section{Introduction}

LLM agents are increasingly evaluated on long-horizon decision tasks: running a
vending machine for months \citep{vendingbench2025}, managing a retail
store \citep{retailbench2026}, operating supply chains
\citep{aimbench2025,invagent2024}. These tasks share a
structure. The world has hidden state, such as a demand shift or a failing
supplier, that the agent can only infer from noisy symptoms, and performance
depends on acting on those inferences week after week. Benchmarks in this genre
consistently report that agent performance degrades over long horizons. What
they largely cannot report is \emph{why}.

When an agent incurs excess cost on such a task, two distinct failure modes
are possible. It may have failed to \emph{infer} the hidden state, never
recognizing that demand had shifted; or it may have inferred correctly and
failed to \emph{act}, stating the correct diagnosis in its reasoning while
placing the wrong order.
The second failure mode is documented in bandit and game settings under the
name of the \textbf{knowing-doing gap}
\citep{schmied2025greedy,sobotka2026broken}: models often hold or state
correct beliefs about the task yet fail to act on them.
These are failures of perception and failures of action --- in the language
of partially observable decision processes, failures of \emph{state
estimation} and failures of \emph{control} --- and they call for
different fixes: better inference scaffolding in the first case, better
policy elicitation in the second. Telling them apart, however, requires knowing
what a correct inference \emph{was worth}.

Existing evidence for the knowing-doing gap does not support that accounting.
It comes from probes and single decisions in settings where the correct
action is known in closed form \citep{sobotka2026broken,schmied2025greedy},
or from cost comparisons against a reference policy that sees privileged
information: the oracle in RetailBench, for instance, is computed with
access to ground truth the agent never observes \citep{retailbench2026}. A
privileged oracle conflates the two failure modes by construction, because the
shortfall it measures bundles ``did not know'' and ``knew but did not act''
into one number. To attribute cost to the knowing-doing gap specifically, the
reference policy must be denied exactly the information the agent is denied.

We introduce a benchmark in which that fair reference is computable. The task
is a 26-week inventory-replenishment problem, formalized as a factored POMDP
with six hidden factor processes whose states the agent must infer from noisy
weekly symptoms (Section~3). Because the hidden structure is factored and the
event tape fixed, an explicit Bayesian reference policy is feasible:
an exact Bayes filter per factor driving a rollout policy, conditioned on
exactly the observation stream the LLM receives. This reference is fair rather than
privileged; Section~4 states the one disclosed asymmetry. Together with a
symptom-blind base-stock floor, it yields a \textbf{skill score} locating
each agent between ignoring the symptoms (0) and matching the fair Bayesian
(1), and two rationale-based metrics, \textbf{detection lag} and the
\textbf{knowing-doing rate}, that separate state estimation from control.
Fifty seeds, grouped by stress profile (\textsc{isolated},
\textsc{persistent}, \textsc{compound}), make the interaction of hidden
failures a controlled variable.

Our contributions are:
\begin{enumerate}
  \item We introduce a six-factor supply-chain POMDP benchmark whose upper
  reference is a fair, computable Bayes-filter oracle: an exact-filter
  rollout policy conditioned on the identical observation stream as the
  agent, referenced as the mean of 20 replications per seed. Fairness is
  what makes agent shortfall attributable to acting rather than confounded
  with knowing.
  \item We construct a controlled stress axis (\textsc{isolated} /
  \textsc{persistent} / \textsc{compound}) that varies whether and how
  hidden failures overlap while holding the world model fixed.
  \item We measure state estimation and control separately --- skill
  score, detection lag, knowing-doing rate --- over 200 episodes:
  Claude Sonnet 5, GPT-5.4, DeepSeek-V4-Pro, and Grok 4.5, each on all
  fifty seeds, under a rules-only prompt with no strategy playbook.
\end{enumerate}

Across 200 episodes, detection proves nearly uniform: all four models name
84--88\% of hidden-factor episodes, typically within a week of onset, and
the two models with the lowest skill scores name them fastest. Costs
diverge sharply nonetheless: two of the four models finish below the
symptom-blind floor overall --- on roughly half their seeds, a rule that
reads no symptoms at all would have been cheaper than acting on their
largely correct diagnoses. The gap also concentrates: on
\textsc{persistent} seeds, 34--43\% of correctly diagnosed stress weeks
still end in stockout for every model, a rate Section~5 shows is partly
severity (weeks bad enough to notice are also bad enough to strain the
shelf). What separates models is the distance between seeing a problem and
acting on it, and the failures run in both directions: ordering too little
too late, and paying too much for protection.

\begin{figure}[t]
  \centering
  \includegraphics[width=\textwidth]{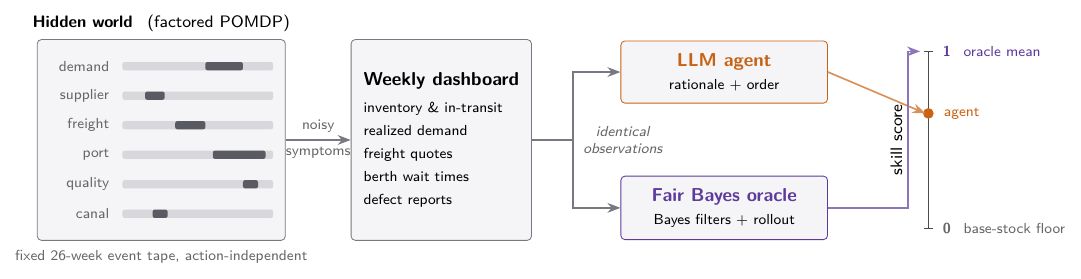}
  \caption{\textbf{Measuring the knowing-doing gap with a fair oracle.} Six
  hidden factor processes evolve on a fixed, action-independent event tape
  (left; dark segments mark stress regimes) and emit noisy symptoms into a
  weekly dashboard. The identical observation stream feeds both the LLM agent
  and a Bayes-filter reference policy (one exact filter per factor driving a
  rollout), so neither sees the hidden state. The agent's episode cost is
  placed on a skill scale anchored by a symptom-blind base-stock floor (0)
  and the fair oracle's 20-replication mean (1): shortfall on this scale is
  attributable to acting on beliefs, not to forming them.}
  \label{fig:overview}
\end{figure}

\section{Related Work}
\label{sec:related}

\begin{table}[t]
  \caption{The benchmarks closest to STOCKTAKE. Cells condense each paper's
  own description of its task and reference policies; citations in text.
  RetailBench builds its simulator from real retail records; ours is
  synthetic and seeded.}
  \label{tab:related}
  \centering
  \small
  \setlength{\tabcolsep}{5pt}
  \begin{tabular}{@{}l >{\raggedright\arraybackslash}p{2.0cm} >{\raggedright\arraybackslash}p{1.6cm} >{\raggedright\arraybackslash}p{2.9cm} >{\raggedright\arraybackslash}p{3.1cm} l@{}}
    \toprule
    Benchmark & Setting & Horizon & Hidden state & Reference policy & Belief graded \\
    \midrule
    Vending-Bench & vending machine & 2{,}000 messages & per-step noise only & none (human baseline) & no \\
    AIM-Bench & five inventory envs & 20 rounds (NVP); varies & stochastic demand, lead times & hindsight optimum & no \\
    RetailBench & retail store sim & 180 days & news-impact and supplier params & rule-based, privileged & no \\
    \midrule
    \textbf{STOCKTAKE} & import replenishment & 26 weeks & six persistent factor processes & Bayes-filter rollout on the agent's observations & yes \\
    \bottomrule
  \end{tabular}
\end{table}

Evaluating LLM agents as sequential decision-makers has grown from
single-turn tool use into month-scale simulations of running a business.
Vending-Bench \citep{vendingbench2025} has agents operate a vending machine
across two thousand messages and finds coherence collapse, ruling out
context exhaustion as the cause. RetailBench \citep{retailbench2026}
simulates a supermarket built on real retail records, frames it as a
partially observable decision process, and finds that only a few models
survive its 180-day horizon. In inventory management specifically,
AIM-Bench \citep{aimbench2025} finds human-like ordering biases
(demand anchoring, bullwhip amplification) across five
environments, and InvAgent \citep{invagent2024} reports outcome gaps
against heuristic and reinforcement-learning baselines in a fully observed
twelve-period setting.

A second line of work asks whether agents act on what they know. In bandit
tasks, models produce a correct rationale for 87\% of decisions and still
take the greedy action over the optimal action they just derived
\citep{schmied2025greedy}. Probing studies in strategic games find internal
beliefs about the opponent that are often accurate, weakly verbalized, and
weakly linked to the chosen action \citep{sobotka2026broken}. Probes of
tool-use decisions find the same split inside the network: the belief that
an action is needed and the action taken are both linearly decodable, but
their directions drift apart in later layers \citep{cheng2026toolgap}. The
phenomenon has a name, the knowing-doing gap, and it has so far been
measured on decisions whose correct answer is known in closed form.

A third line evaluates the beliefs themselves. BayesBench
\citep{bayesbench2026} scores posterior updating against a rational Bayesian
reasoner and finds systematic departures, on estimation tasks with no
actions attached. The machinery for connecting such beliefs to inventory
decisions is classical: Bayraktar and Ludkovski
\citep{bayraktar2010inventory} derive the exact Bayes-filter reduction of
inventory control with
hidden Markov-modulated demand to a fully observed problem, in their case
to compute optimal policies.

The task is causal only in a specific, limited sense, and we place it
against that literature explicitly. Causal-reasoning benchmarks test
whether a model can recover causal structure or answer interventional and
counterfactual queries \citep{cladder2023}, and there is evidence that
models recite causal facts from text rather than infer them
\citep{zecevic2023parrots}. Our task tests neither: the causal structure
of the world is given to the agent in the prompt, and actions never touch
the hidden dynamics, so there is nothing to discover and nothing to
intervene on. What the weekly decision demands is abduction, inferring
which hidden cause best explains the week's noisy symptoms under a known
model, followed by cost-sensitive control. The benchmark is
complementary to causal-reasoning evaluations: it measures whether a
correct causal attribution, once formed, survives the trip to an action.

Table~\ref{tab:related} summarizes the benchmarks closest to ours. In each,
the reference policy either reads information the agent is denied
(RetailBench's rule-based oracle takes ``structured simulator fields'' as
input; AIM-Bench's ex-post optimum is computed from realized demand) or is
absent beyond a single human run, and none grades the agent's stated
beliefs alongside its actions. STOCKTAKE addresses both limitations with
one construction: the classical filter reduction, repurposed as a reference
policy conditioned on exactly the agent's observation stream, so that
shortfall against it isolates the action side of the knowing-doing gap,
while the weekly stated beliefs it is compared against are graded on the
same episodes.

\section{STOCKTAKE: Task and Environment}

STOCKTAKE casts the agent as the replenishment manager of an electronics
importer for a 26-week episode. The state that drives cost is never
observed directly. The agent reads weekly symptoms, commits an
order, and writes down what it believes is happening
(Figure~\ref{fig:overview}).

\subsection{The weekly decision}

Each week the agent reads a dashboard of on-hand inventory, orders in
transit, realized demand, freight quotes, berth waiting times, canal
transit counts, supplier scorecards, and quality reports. It may then take
any of five within-week actions, none of which advances time.
\texttt{lock\_freight} fixes the current freight multiplier for a chosen
number of weeks; the lock itself is free, but a locked rate forgoes any
drop in the spot index. \texttt{expedite\_air} flies up to 20 units in at
\$15 per unit; they land the following week and bypass the port entirely.
\texttt{inspect\_batch} (\$40) sorts an arriving batch so that most
defective units are recovered before they stock. \texttt{buy\_briefing}
(\$30) and \texttt{buy\_audit} (\$25) reveal the true state of the canal
and of the spot supplier, respectively.

The week is then committed by exactly one call to
\texttt{place\_order(rationale, qty, route, supplier, contract)}. The
order names a quantity, a route (through the Suez Canal, or slower and
dearer around the Cape), and a supplier; the optional contract action
signs, switches, renews, or lapses a supplier contract on short, long,
strict, or lenient terms, and the agent can only source suppliers it holds
a live contract with. The \texttt{rationale} argument is required, and the
world does not advance without it. Every run therefore contains 26 written
statements of what the agent believed each week; Section~4 derives the
belief-side metrics from them.

Costs accrue for units held and in transit, for demand missed, and for the
priced levers above; the episode score is total cost, lower is better. The
price ladder is the task's core tension: sea freight is \$4 per unit and
takes three weeks, air is \$15, and every unit of missed demand costs
\$20. An agent that reads the symptoms early can buy its way out of
trouble cheaply; one that reads them late cannot.

\subsection{The hidden world}

\begin{figure}[t]
  \centering
  \includegraphics[width=\textwidth]{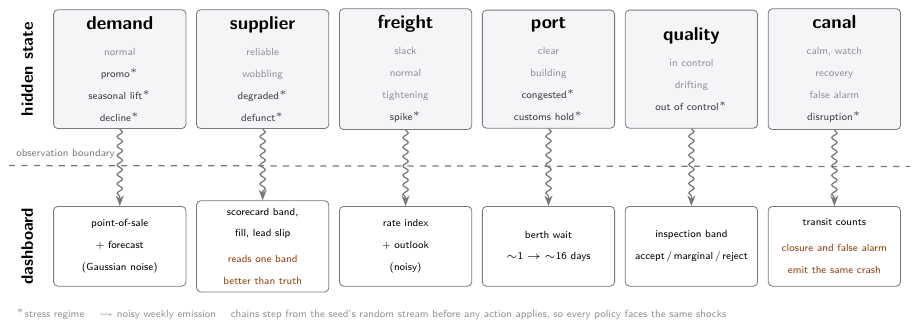}
  \caption{The hidden world behind the dashboard. The hidden state is six
  independent Markov chains; starred regimes are the stress regimes that
  define episodes in Section~4. Each chain emits one symptom stream
  into the weekly dashboard, and those streams are all the agent (and the
  oracle) ever sees. Two symptoms mislead by construction (orange): the
  supplier scorecard reads one severity band better than the truth, and
  the canal's transit counts crash identically for a real closure and a
  false alarm.}
  \label{fig:world}
\end{figure}

The environment is a factored POMDP. Two structural properties of its
hidden state make a fair reference policy computable. Let $x_t = (x_t^1,
\dots, x_t^6)$ denote the hidden state in week $t$, one factor process
each for demand regime, supplier health, freight market, port congestion,
batch quality, and canal status, each a small Markov chain over
regime--age pairs that moves between a baseline and one or more stress
regimes (Figure~\ref{fig:world}). The regime and its age set the
transition probabilities: how likely a stress episode is to start, to
persist once started, and to resolve.

First, \textbf{the dynamics are factored and action-independent}. Each
factor steps as $x_{t+1}^i \sim P^i(\cdot \mid x_t^i)$, reading neither
the other factors nor the agent's action. Algorithm~\ref{alg:tape} is
the entire generative process: one random stream, seeded once, consumed
by the six factors in a fixed order. Because no step reads an action,
the seed fixes every regime change and every noisy reading for all 26
weeks before the agent exists; we call this fixed record the episode's
\emph{event tape}. Two policies run on the same seed face byte-identical
shocks and symptom noise, so the reference policies of Section~4 are
evaluated on the very tape the agent lived through, not on a re-rolled
one. Actions decide how much each shock costs, never what happens in the
hidden world.

\begin{algorithm}[t]
\caption{Event-tape generation. One random stream $r$, seeded once, is
consumed by the six factors in a fixed order; no step reads an agent
action, so the tape is a pure function of the seed. Lines 5--10 are the
transition kernel $P^i$ and line 11 the emission kernel $Q^i$ of
Section~3.2; every onset rate, persistence rate, age cap, and band mean
is tabulated in Appendix~\ref{app:world}.}
\label{alg:tape}
\begin{algorithmic}[1]
\State $r \gets \textsc{Random}(\mathit{seed})$
\State $(s^i, a^i) \gets (\text{baseline}^i,\, 0)$ for each factor $i$ \Comment{regime and its age}
\For{week $t = 1, \dots, 26$}
  \For{factor $i$ in fixed order} \Comment{canal, supplier, demand, freight, port, quality}
    \If{$s^i$ is the baseline}
      \State leave it for regime $k$ w.p.\ $\mathrm{onset}^i_k$, else stay \Comment{e.g.\ clear $\to$ building}
    \Else
      \State stay w.p.\ $\mathrm{persist}^i(s^i, a^i)$, else advance a stage or recover \Comment{forced out at $\mathrm{cap}^i(s^i)$}
    \EndIf
    \State $a^i \gets a^i{+}1$ if the regime held, else $0$; \quad $x_t^i \gets (s^i, a^i)$
    \State $y_t^i \gets$ factor $i$'s readings, drawn from $r$ around its band's mean \Comment{e.g.\ $\mathrm{round}\,\mathcal{N}(\mu_{\text{band}}, \sigma)$}
  \EndFor
\EndFor
\State \Return the tape $(x_{1:26},\, y_{1:26})$
\end{algorithmic}
\end{algorithm}

The port factor instantiates the template. From clear, the port can
start building toward a backlog or take a sudden customs hold (line~6);
building tips into congested or clears (line~8); congested persists week
to week until its age cap forces recovery. Each week the realized berth
wait is drawn around the current band's mean, roughly one day when clear
and sixteen when congested, and while congested every arriving shipment
is held an extra week. The band on line~11 is where the benchmark's
ambiguities are constructed: for several factors the first week of two
different regimes maps to the same band, so a fresh customs hold and
newly tipped congestion both read ``slow'', and a demand promotion and a
seasonal lift both open on the same ``surge''. A single reading cannot
say which stress has begun; only the following weeks separate them. A
few factors draw auxiliary state from the same stream in the same fixed
order, such as the canal event's type at onset.

Second, \textbf{the observations are factored}. The weekly dashboard
$o_t$ contains the fully observed logistics state (on-hand stock, orders
in transit, realized demand) and one symptom channel per factor,
$y_t^i \sim Q^i(\cdot \mid x_t^i)$. The agent never observes any $x_t^i$
directly; the paid briefing and audit of Section~3.1 are the only
exceptions, each revealing one factor's current state for a fee. The
channels themselves are hard to read in three different ways. Most are
noisy: a building port and a clear one differ by a few days of berth
wait, against scatter of the same size. The supplier scorecard is
biased, reading one severity band better than the truth by design. The
canal channel is exact but ambiguous, emitting the
same transit-count crash for a false alarm as for the first week of a real
closure, so even a Bayesian observer \emph{should} be uncertain there.

The cost function, by contrast, is \emph{not} factored. The factors
evolve independently but bill jointly through a single set of books: a
port hold delays the same shipments a demand surge is waiting on, and a
quality escape shrinks the same inventory that surge is draining.
Concurrent shocks can therefore cost far more than the same shocks
arriving in sequence, and Section~3.3 samples that regime deliberately.
Full cost constants, caps, lead times, and per-factor transition
parameters are in Appendix~\ref{app:world}.

\subsection{Seeds and stress profiles}

Because the tape is a pure function of the seed
(Algorithm~\ref{alg:tape}), instances can be minted and classified
before any agent exists. Any unsigned integer names a candidate: the
world is replayed to week 26 with empty actions, and the resulting tape
is summarized by a handful of features --- the number of stressed weeks
per factor, how many weeks have two or more factors stressed at once and
the most stressed simultaneously, the longest unbroken run of port
blockage, and whether the spot supplier failed outright. Thresholds on
these features classify the candidate's stress profile, and candidates
too benign to separate good decision-making from bad (those on which the
two reference policies of Section~4 cost nearly the same) are discarded.
The benchmark fixes fifty seeds chosen this way from a 200-seed pool:
\begin{itemize}
  \item \textsc{isolated} (12 seeds), short, non-overlapping
        single-factor shocks;
  \item \textsc{persistent} (15 seeds), a multi-week stretch of port
        congestion overlapping a demand rise. Here the intuitively
        appealing response, freezing orders until the congestion clears,
        is the costly one, because orders placed \emph{through} the
        congestion arrive as it clears;
  \item \textsc{compound} (23 seeds), pileups of three or more concurrent
        factor episodes.
\end{itemize}
Compound stress is deliberately oversampled relative to the pool; it is the
regime the benchmark targets, and all results are reported per group.
Because membership is fixed by the tape alone, the structure of hidden
stress is an independent variable of the benchmark, and the same minting
recipe extends it: new instances of any profile can be generated and
classified with no agent runs and no human labeling.

\subsection{Agent prompt}

Agents run under a single \emph{rules-only} prompt that specifies the
mechanics of the task (costs, lead times, levers, output format) and gives
a qualitative description of the world's factors, with no strategy advice,
no playbook, and no worked examples. The benchmark thereby measures what a
capable agent brings to the task on its own. The full
prompt and the weekly observation format are reproduced in
Appendix~\ref{app:prompt}.

\section{Experimental Setup}

\textbf{Reference policies.} Every run is scored between two reference
policies evaluated on the same seed, and both are functions of the agent's
own observation stream $o_{1:t}$, nothing more. The floor is a textbook
base-stock policy: order up to a fixed target $S = \mu L +
z\sigma\sqrt{L}$ (mean demand $\mu$, lead time $L$, safety factor $z$ from
the holding/stockout cost ratio) every week, always by the same route and
supplier, reading no symptom channel $y_t^i$ at all. The upper reference
is the Bayes-filter oracle of Figure~\ref{fig:overview}. Because the
dynamics and the observations both factor and neither reads the action
(Section~3.2), the posterior over the hidden state splits into six
independent parts, each small enough to maintain \emph{exactly}: one
forward (predict--correct) filter per factor tracks
$b_t^i = \Pr(x_t^i \mid y_{1:t}^i)$, updated each week on the same
observation dictionary $o_t$ that is serialized into the LLM prompt. To act, the oracle samples $\sim$200 possible futures of
all six factors from the belief $b_t = (b_t^1, \dots, b_t^6)$, scores a
small menu of candidate actions (order sizes, routes, each mitigation
lever, gated on the relevant posterior risk) against every sampled future,
and executes the cheapest candidate through the same action API the LLM
uses. Its randomness is independent of the world's, and it never reads
hidden state. One asymmetry is disclosed: its filters use the
environment's true generative parameters (the transition tables $P^i$ and
the regime means behind $Q^i$), which the LLM receives only as a
qualitative description. We
therefore interpret the oracle as Bayes-optimal given the true model.
Because it is a stochastic policy, its per-seed reference value is the mean
of 20 replications (standard error 0.2--1.8\% of the reference cost); it is
a fair reference rather than an optimum, and agents exceed it on some tapes
(Section~5).
Appendix~\ref{app:oracle} specifies both policies in full.

\textbf{Metrics.} Let $C_{\mathrm{base}}$, $C_{\mathrm{orc}}$, and
$C_{\mathrm{agent}}$ denote total episode cost for the floor, the oracle
reference, and the agent. The skill score
\begin{equation}
\mathrm{skill} \;=\; \frac{C_{\mathrm{base}} - C_{\mathrm{agent}}}{C_{\mathrm{base}} - C_{\mathrm{orc}}}
\end{equation}
is 0 at the symptom-blind floor and 1 at the oracle mean; values above 1 and
below 0 both occur. One of the fifty seeds has negative headroom
($C_{\mathrm{base}} < C_{\mathrm{orc}}$): the floor already beats the oracle
mean on that tape, so skill is undefined there, and the seed is excluded
from every skill mean (leaving $n = 49$). Its runs still count in a
denominator-free companion statistic, the number of seeds on which the
agent's raw cost beats the floor's. Three further seeds have headroom under
\$700, where the ratio is noisy; they are flagged in the per-seed results
but not excluded. A worked example from seed~1: base-stock costs \$9{,}101,
the oracle mean is \$6{,}932, and Claude Sonnet 5 spent \$8{,}539, so its
skill is $(9101-8539)/(9101-6932)=0.26$ --- it recovered about a quarter of
what seeing the symptoms was worth on that tape. For belief-side metrics, a
\emph{stress episode} is a maximal run of consecutive weeks in which one
factor is truly in a stress regime (283 episodes per model across the
fifty tapes).
Detection lag is the number of weeks from episode onset until the factor is
first named in the agent's rationale (averaged over detected episodes;
never-detected episodes are reported separately), and the knowing-doing rate
is
\begin{equation}
\mathrm{KD} \;=\; \Pr\!\left(\,\text{stockout in week } t \;\middle|\; \text{a truly stressed factor is named in week } t\,\right),
\end{equation}
estimated over the group's correctly diagnosed stress weeks, where
``stockout'' means the stock available that week (inventory plus arrivals)
fell short of realized demand. A high KD rate means correct diagnosis coexisting with empty
shelves.

\textbf{Belief grading.} Whether a rationale names a factor is judged by a
small LLM grader (gpt-5-mini, temperature 0) that sees only the rationale
text and the fixed factor vocabulary, never the tape; labels are aligned to
the week the rationale was written. Grader labels were audited in two rounds
against raw traces; residual disagreements are borderline over-labels on calm
weeks, which cannot enter the knowing-doing rate by construction. The rationale is stated
belief: a lower bound on what the model knows, a point we return to in
Section~6. Grading rules and validation details are in
Appendix~\ref{app:grading}.

\textbf{Models.} We evaluate Claude Sonnet 5, GPT-5.4, DeepSeek-V4-Pro,
and Grok 4.5, each on all fifty seeds. All four run through an identical
harness (OpenRouter model IDs \texttt{anthropic/claude-sonnet-5},
\texttt{openai/gpt-5.4}, \texttt{deepseek/deepseek-v4-pro},
\texttt{x-ai/grok-4.5}; temperature 0; runs completed July 2026), one run
per model--seed pair, 200 episodes in total, each validated to the full 26
weeks, all under the rules-only prompt of
Section~3.\footnote{An earlier circulated draft reported DeepSeek-V4-Pro
on a twenty-seed subset: a fault in the belief-grading pipeline had
silently read stale run files for the remaining seeds. All numbers here
are from complete fifty-seed runs, regraded and rescored.}

\section{Results}

\begin{figure}[t]
  \centering
  \includegraphics[width=\textwidth]{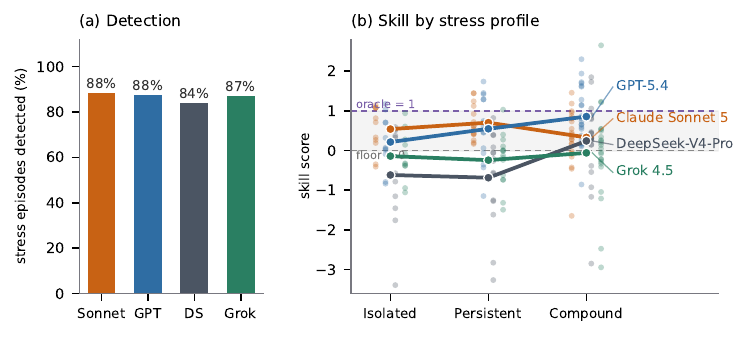}
  \caption{\textbf{Seeing is uniform; doing is not.} (a) All four models
  detect 84--88\% of their hidden-factor stress episodes (283 per model
  over the fifty seeds), typically within a week of onset. (b) Group-mean
  skill by stress profile (lines), with per-seed scores as faint dots.
  Costs diverge across models despite panel (a): DeepSeek-V4-Pro and
  Grok 4.5 finish below the symptom-blind floor overall.}
  \label{fig:results}
\end{figure}

\begin{table}[t]
  \caption{Skill score by stress profile: group mean (median), single run
  per model--seed pair, every model on all fifty seeds (11/15/23 scoreable
  per group after the seed with negative headroom is excluded). \emph{neg}
  counts scoreable seeds below the base-stock floor; \emph{beats floor}
  counts every seed, including the excluded one, on which the agent's raw
  cost beat the floor's.}
  \label{tab:skill}
  \centering
  \small
  \begin{tabular}{lcccccc}
    \toprule
    Model & \textsc{isolated} & \textsc{persistent} & \textsc{compound} & All & neg & beats floor \\
    \midrule
    Claude Sonnet 5 & 0.54 (0.71) & 0.70 (0.61) & 0.33 (0.35) & 0.49 (0.45) & 9 & 40/50 \\
    GPT-5.4         & 0.21 (0.07) & 0.55 (0.61) & 0.86 (0.73) & 0.62 (0.65) & 8 & 42/50 \\
    DeepSeek-V4-Pro & $-$0.62 ($-$0.57) & $-$0.68 ($-$0.26) & 0.24 (0.32) & $-$0.23 (0.15) & 22 & 28/50 \\
    Grok 4.5        & $-$0.14 ($-$0.32) & $-$0.24 ($-$0.12) & $-$0.06 (0.05) & $-$0.13 ($-$0.08) & 28 & 21/50 \\
    \bottomrule
  \end{tabular}
\end{table}

\begin{table}[t]
  \caption{Belief-side metrics. Detection is pooled over each model's
  stress episodes; the knowing-doing rate is the fraction of correctly
  diagnosed stress weeks that still ended in stockout, pooled over the
  group's stress weeks. Table~\ref{tab:confound} reports the severity
  confound that qualifies these rates.}
  \label{tab:belief}
  \centering
  \begin{tabular}{lcccccc}
    \toprule
    & \multicolumn{2}{c}{Detection} & \multicolumn{4}{c}{Knowing-doing rate} \\
    \cmidrule(lr){2-3}\cmidrule(lr){4-7}
    Model & missed & mean lag (wk) & \textsc{isol.} & \textsc{persist.} & \textsc{comp.} & all \\
    \midrule
    Claude Sonnet 5 & 33/283 & 0.42 & 0.24 & 0.43 & 0.30 & 0.34 \\
    GPT-5.4         & 35/283 & 0.40 & 0.08 & 0.37 & 0.22 & 0.26 \\
    DeepSeek-V4-Pro & 46/283 & 0.32 & 0.07 & 0.40 & 0.22 & 0.26 \\
    Grok 4.5        & 37/283 & 0.32 & 0.09 & 0.34 & 0.21 & 0.24 \\
    \bottomrule
  \end{tabular}
\end{table}

\begin{table}[t]
  \caption{Severity confound check for the knowing-doing rate: stockout
  rate on truly stressed weeks, split by whether the week's rationale
  correctly diagnosed the stress (week counts in parentheses). Diagnosed
  weeks stock out \emph{more} than undiagnosed ones for every model, so
  part of the knowing-doing rate reflects severity rather than inaction.}
  \label{tab:confound}
  \centering
  \begin{tabular}{lcc}
    \toprule
    Model & diagnosed stress weeks & undiagnosed stress weeks \\
    \midrule
    Claude Sonnet 5 & 0.34 (515) & 0.19 (162) \\
    GPT-5.4         & 0.26 (502) & 0.14 (175) \\
    DeepSeek-V4-Pro & 0.26 (469) & 0.11 (208) \\
    Grok 4.5        & 0.24 (544) & 0.12 (133) \\
    \bottomrule
  \end{tabular}
\end{table}

All four models see the hidden world about equally well
(Table~\ref{tab:belief}, Figure~\ref{fig:results}a). Each detects 84--88\% of
its stress episodes, with a mean lag of 0.32--0.42 weeks: when a factor
enters a stress regime, its symptoms are typically named in the rationale
within the same week. The two models that end up below the base-stock
floor on cost have the \emph{shortest} detection lags. Whatever separates
these models on this task, it is not state estimation.

Skill scores, in contrast, diverge widely (Table~\ref{tab:skill},
Figure~\ref{fig:results}b). Claude Sonnet 5 and GPT-5.4 recover 49\% and
62\% of the floor-to-oracle cost gap overall and beat the base-stock
floor's raw cost on 40 and 42 of 50 seeds. DeepSeek-V4-Pro and Grok 4.5
score $-0.23$ and $-0.13$, falling below the floor on 22 and 28 of their
49 scoreable seeds: on roughly half their tapes, a symptom-blind
base-stock rule with the same action menu would have been cheaper than
acting on their (largely correct) diagnoses. Along the stress axis the
field splits three against one. GPT-5.4, DeepSeek-V4-Pro, and Grok 4.5
all post their best group on \textsc{compound} seeds (GPT-5.4 strongest
at 0.86, with its weakest group \textsc{isolated} at mean 0.21, median
0.07), while Claude Sonnet 5 is the exception: strongest on
\textsc{persistent} (0.70, no seed below the floor) and weakest on
\textsc{compound}, where seven of its nine below-floor seeds sit.
DeepSeek's \textsc{compound} mean is pulled up by a single outlier tape;
its group median is 0.32. Individual scores above 1 occur (33 of 196
scoreable runs) and are legitimate, since the oracle is Bayes-optimal in
expectation but not per-tape. Even the best model leaves over a third of
the gap to a policy computed from identical observations, so the benchmark
is far from saturated.

The knowing-doing rate (Table~\ref{tab:belief}) shows where correct diagnosis
fails to prevent empty shelves. For every model the rate peaks on
\textsc{persistent} seeds, at 0.34--0.43, versus 0.07--0.24 on
\textsc{isolated} seeds: when a disruption lasts many weeks, a correctly
diagnosed stress week still ends in stockout roughly four times in ten.
Brief shocks are handled; sustained ones convert knowledge into cost.
The rate is not a model ranking, however: read across models it runs
\emph{opposite} to skill, with the near-best model (Claude Sonnet 5)
highest and the worst (Grok 4.5) lowest. The reason is that a
stockout-based rate sees only one face of the gap. An agent can fail on a
correct diagnosis by under-responding, stocking out despite naming the
problem, or by over-responding, buying so much premium protection that
stockouts are rare while the bill exceeds what the symptom-blind floor
would have paid. The two models below the floor fit the second face:
their diagnosed-week stockout rates are the lowest of the four while
their total costs are the highest. We quantify the response taxonomy only
for the two above-floor models (below); for the two below it, the
mechanism rests on trace readings like Exhibit~B in Appendix~D, not on a
measured cost decomposition. Two further qualifications keep
the rate from reading as a week-level fault measure. A
stockout in a given week can stem from an ordering decision made weeks
earlier. And Table~\ref{tab:confound} shows a severity confound: stressed
weeks the agent correctly diagnosed stock out more often than stressed
weeks it missed, for every model, because weeks bad enough to be noticed
are also bad enough to strain the shelf. The knowing-doing rate therefore
says that correct diagnosis does not guarantee protection; it does not say
that diagnosis causes stockouts, and we do not rank models on it.

Reading the traces suggests the models fail in opposite directions; full
excerpts are in Appendix~\ref{app:seeds}. Claude Sonnet 5 on seed~1
(\textsc{persistent}; port congested weeks 17--24 while demand lifts)
\emph{freezes}: it diagnoses the congestion in week~16 and skips its sea
order at once; the orders it does place are held up behind the blocked
port, and through weeks 22--24, at zero inventory, it stops ordering by sea
entirely (Figure~\ref{fig:seed1} in the appendix plots the full episode).
By week~22 it has held zero inventory for two weeks, yet writes
``I already have a very large pipeline: 104 units in transit \ldots\ ordering
more now would only add to an already oversized pipeline \ldots\ given
berth\_wait is still elevated (15 days)'' and orders nothing, counting units
trapped behind the congestion it itself diagnosed as usable cover. It ends
with four consecutive zero-inventory weeks costing \$660--\$823 each, while
the oracle keeps placing modest orders that arrive after the congestion
clears. DeepSeek-V4-Pro on seed~25 (\textsc{isolated}; a two-week canal
closure) \emph{flails}: within two weeks of a correct diagnosis it locks
freight rates for four weeks, air-expedites units, and stacks 80 units of
orders against $\sim$20-per-week demand, on top of 117 already in
transit. The closure ends in week~5; the fees and holding costs it bought
persist, and the episode finishes at skill $-1.45$; the costly response
outlived the two-week shock it answered. Neither trace contains a
detection error. The two directions generalize beyond these seeds. In
every one of Claude Sonnet 5's nine below-floor seeds, its spending on
air expedite alone exceeds its entire excess over the floor, and about
nine in ten of its air orders on those seeds coincide with a genuine
disruption: it over-doses real problems rather than chasing noise.
GPT-5.4 errs the other way, structurally under-using the mitigation
levers: it never locks freight rates in 87\% of the seeds containing a
freight spike and never orders a batch inspection in 63\% of the seeds
with quality escapes, while naming those conditions in its rationales.

\section{Discussion and Limitations}
\label{sec:discussion}

The results support one claim: on this task, frontier-model failures are
concentrated in control, not state estimation, and the fair oracle makes that
split measurable. A privileged oracle could not have licensed it, because any
shortfall could be attributed to the information gap rather than to policy.
The fairness itself carries one disclosed asymmetry. The oracle receives the
identical observation stream and never the hidden state, but its filters use
the true generative parameters (regime means, transition probabilities),
while the LLM sees only a qualitative world description. Skill~$=1$
therefore means ``Bayes-optimal given the true model'', a demanding but not
information-free reference, and scores above 1 are possible and observed.

The belief metrics read stated beliefs, and stated belief is a lower bound.
Detection is measured from the rationale the model writes, not from its
internal state; a model may track a factor it never mentions. This biases
detection down, which strengthens the seeing-vs-doing contrast, but the
knowing-doing rate inherits the grader's labels. A stratified manual read
found the labels correct on the sampled metric-relevant weeks, and an
independent second grader agrees with the production grader on the
metric-relevant component (which named factors intersect true stress) at
89\%; residual disagreement sits on calm-week over-labels, which cannot
enter the knowing-doing rate by construction. Both audits sampled the
original twenty-seed grading run; all later grading (the thirty expansion
seeds and the additional models) used the identical configuration but was
not freshly audited.

Two design choices qualify behavioral readings of the traces. Once a
second supplier contract is signed, the world offers no way to end it
(the lapse action affects only unsigned offers), so the weekly
dual-sourcing fee is unavoidable from that point on, for every agent, and
paying it is not by itself a decision error. And the two paid
intelligence levers (audit and briefing) return near-deterministic ground
truth by design, so what varies across models is how often they are
consulted, not how well their output is read.

The cost-side numbers rest on single runs. Each model--seed cell is one
run, and provider-side nondeterminism is unquantified. Group means pool
11--23 seeds, and the cross-model contrasts (a 0.85-point overall skill
spread against detection rates within four
points of one another) are far larger than plausible single-run noise, but
per-seed values should be read as samples, not estimates. The oracle
reference is a 20-replication mean with standard error 0.2--1.8\% of the
reference cost.

Finally, the benchmark covers one domain, one SKU, one prompt arm, and
four models. The tapes are generated from published seeds rather than
scraped, so they cannot occur in training data, but conclusions about
\emph{why} the gap concentrates on persistent stress await intervention:
prompt arms that coach explicit belief maintenance, repeated runs per cell,
a clairvoyant ceiling to bound the oracle's optimality gap, and a broader
model set.

\newpage
\appendix
\section*{Appendix}

\section{World and Cost Parameters}
\label{app:world}

All values below are read directly from the environment source; they fully
specify the weekly cost function and the six hidden factor processes of
Section~3.

\paragraph{Costs.} Each week's cost is the sum of the applicable components:

\begin{center}
\small
\begin{tabular}{lll}
\toprule
Component & Charge & When \\
\midrule
Sea shipping & \$4.00 (Suez) / \$6.00 (Cape) per unit $\times$ freight mult. & at dispatch \\
Supplier price delta & qualified $+$\$1.00, spot $-$\$1.50, backup $+$\$0.30 per unit & at dispatch \\
Holding & \$1.00 per unit-week, on hand and in transit & weekly \\
Stockout & \$20.00 per unit of unmet demand (lost sales) & weekly \\
Port demurrage & \$2.00 per unit held at a blocked port & while blocked \\
Quality rework & \$15.00 per defective unit landed & at arrival \\
Diversion surcharge & \$2.00 per unit forced from Suez around the Cape & at diversion \\
Crisis back-order & \$60.00 per unit a supplier shorts during a canal event & coupled \\
Air expedite & \$15.00 per unit, at most 20 units/week & on use \\
Briefing / audit / inspection & \$30 / \$25 / \$40 flat per call & on use \\
Dual-sourcing overhead & \$4.00 per week while two or more contracts are live & weekly \\
\bottomrule
\end{tabular}
\end{center}

Orders are capped at 100 units per week. Sea lead times are 3 weeks via
Suez and 4 via the Cape; air freight lands the following week. The implied
newsvendor critical ratio is $20/21 \approx 0.95$, which fixes the
base-stock safety factor $z \approx 1.67$ used by the floor policy.

\paragraph{Factor processes.} Each factor is an independent Markov chain
stepped from the seed's random stream before any action applies. Starred
regimes are the stress regimes that define episodes in Section~4.

\begin{center}
\small
\begin{tabular}{lp{4.6cm}p{6.2cm}}
\toprule
Factor & Regimes & Key dynamics and observations \\
\midrule
Demand & normal, promo spike$^{*}$, seasonal lift$^{*}$, structural decline$^{*}$ &
onsets 0.03/0.015/0.005 per week; promo lasts 4 weeks, seasonal persists 0.85
(cap 8), decline persists 0.97. Weekly means 20/26/30/14 units; observed
point-of-sale and forecast carry Gaussian noise (sd 4 and 6). \\
Supplier & reliable, wobbling, degraded$^{*}$, defunct$^{*}$ &
onset 0.10, wobbling$\to$degraded 0.45; degraded recovers within 4 weeks or
dies (0.06). Scorecard band reads one severity level better than the truth;
lead-time slip and fill fraction are noisy. \\
Freight & slack, normal, tightening, spike$^{*}$ &
tightening onset 0.06, tightening$\to$spike 0.18, spike persists 0.80 (cap
6). Rate multipliers 0.7/1.0/1.8/4.0; index and outlook noisy. \\
Port & clear, building, congested$^{*}$, customs hold$^{*}$ &
building onset 0.06, building$\to$congested 0.30, congested persists 0.85
(cap 8); customs holds last about a week. Berth wait rises from $\sim$1 to
$\sim$16 days; a blocked port delays every arrival one week. \\
Quality & in control, drifting, out of control$^{*}$ &
drift onset 0.04; the drift-to-failure hazard grows with age
($0.05 + 0.04 \cdot \text{age}$). Defect fractions 0.1\%/2\%/6\%; observed
only as a noisy accept/marginal/reject inspection band. \\
Canal & calm, watch, disruption$^{*}$, recovery, false alarm &
calm$\to$watch 0.08; watch resolves to a real closure (0.50) or a false
alarm (0.20). Transit counts are exact but the first closure week and a
false alarm emit the identical ``crash'' reading. A closed canal queues
Suez ships, then diverts them ($+$3 weeks, Cape rate, surcharge). \\
\bottomrule
\end{tabular}
\end{center}

\section{Reference Policy Details}
\label{app:oracle}

\paragraph{Base-stock floor.} Order-up-to level
$S = \mu L + z\sigma\sqrt{L}$ with $\mu = 20$, $\sigma = 4$, $L = 4$ (3-week
Suez lead plus one review week), $z = \Phi^{-1}(20/21) \approx 1.67$. Every
week it orders $\max(0, S - \text{inventory position})$, always via Suez
from the same supplier, and never touches a lever or reads a symptom.

\paragraph{Oracle decision loop.} Each week the oracle (i) updates one exact
forward Bayes filter per factor on the shared observation dictionary;
(ii) samples $\sim$200 joint futures of all six factors from the filter
posteriors, rolled forward to the episode's final week with the true
transition tables; (iii) scores a
small candidate menu against every sampled future using a fast surrogate of
the cost function, holding future weeks to a base-stock continuation; and
(iv) executes the cheapest candidate through the world's action methods,
the same calls that back the LLM's tools. The candidate menu contains order quantities \{0, four weeks, six
weeks of expected demand\}, both routes, the contracted suppliers, and each
mitigation lever gated on posterior risk (freight lock if
$\Pr(\text{tightening or spike}) > 0.3$; air expedite if port risk $> 0.3$;
inspection if quality risk $> 0.5$ and a batch lands next week). The same
sampled futures are reused across candidates (common random numbers). The
oracle never buys briefings or audits; the LLM may. A seed's reference
value is the mean of 20 such runs; replications vary both the random
stream and the per-decision trajectory budget (191--210 sampled futures,
hence the $\sim$200).

\section{Belief Grading Details}
\label{app:grading}

The grader (gpt-5-mini via OpenRouter, temperature 0, reasoning effort
capped at low) receives only the week's rationale text and the fixed
six-name factor vocabulary, and returns strict JSON. Its instructions require a
factor to be labeled only if the text claims it is a problem \emph{that
week}: ruled-out or benign mentions do not count, factual symptom reports do
count, and precautions without a current problem do not count. Mean
detection lag is computed over detected episodes only; naming a factor
before its episode starts never counts as detection. The search then runs
forward to the end of the trace, so a mention long after onset still
counts, as a slow detection of that episode; in the measured runs mean lag
is under half a week, so detections overwhelmingly land at or just after
onset. In the knowing-doing
rate, ``stockout'' means the week's available stock (inventory plus
arrivals) fell short of realized demand. Validation: on a stratified 30-row
manual read, all 10 sampled hit-class labels were correct and 9 of 10
sampled misses were genuine (the tenth ambiguous, in the direction that
understates detection); an independent second grader (Claude Haiku 4.5, same
prompt, 100 sampled weeks) agreed 43\% on exact label sets but 89\% on the
metric-relevant quantity, the intersection of labeled and truly stressed
factors. Calm-week over-labels, where graders disagree most, are excluded
from every reported metric by construction.

The grader's full instructions are reproduced below; \texttt{\{rationale\}}
is replaced by the week's rationale text.

{\footnotesize
\begin{verbatim}
You will be given one week's replenishment-decision rationale from a supply-
chain manager. The fixed factor vocabulary is exactly these 6 names:
disruption, freight, port, quality, supplier, demand.

Return STRICT JSON only, of the form: {"factors_named": [subset of the 6
names]}

Include a factor ONLY if the rationale asserts it is CURRENTLY
active/problematic this week. Go through the 6 factors one by one and ask:
does the text CLAIM this is a problem right now? Three rules, each with a real
example:

1. RULED-OUT or BENIGN mentions do NOT count. "freight remains cheap" = no
   freight. "the weak fill looks noisy rather than confirmed deterioration" =
   no supplier. "lanes normal" = no disruption. Naming a factor to say it is
   fine is the OPPOSITE of naming it as a problem.
2. FACTUAL symptom reports DO count even without editorializing. "AQL reject
   -- inspected batch" = quality. "port congestion (berth_wait 16) delayed
   arrivals longer than expected" = port. The manager asserts a problem exists
   by reporting its symptom as the cause of something.
3. PRECAUTIONS without a current problem do NOT count. "I'll keep using the
   freight lock" alone = no freight. "avoiding spot until it stabilizes" after
   calling the data noise = no supplier. But acting BECAUSE of an asserted
   current problem ("air-expediting because the port is holding arrivals") =
   count that factor.

Worked example -- rationale: "Freight lock still active (1.19x) shielding us
from the 179 spot index. AQL came back marginal causing a rework charge. Spot
remains slipping (86 OTIF) so still avoiding it." -> {"factors_named":
["freight", "quality", "supplier"]} (index 179 asserted high; AQL/rework is a
current quality symptom; "spot remains slipping" is a current supplier claim.)

Rationale:
"""{rationale}"""

JSON:
\end{verbatim}}

\section{Seed Timelines and Trace Excerpts}
\label{app:seeds}

The table below gives the ground-truth
stress windows for the nine originally released seeds, obtained by replaying
each tape with no actions (the tape is action-independent). Each entry reads
\emph{factor: regime, weeks active}; factors are the six of Section~3.
Windows joined by ``$+$'' overlap in time --- the defining feature of the
\textsc{persistent} and \textsc{compound} groups. Timelines for the
forty-one expansion seeds ship with the released tapes.

\begin{center}
\small
\begin{tabular}{llp{9.2cm}}
\toprule
Seed & Group & True stress windows (weeks active) \\
\midrule
157 & \textsc{isolated} & canal closed 12--13; quality out of control 23--24; freight spike 25--26 \\
25  & \textsc{isolated} & canal closed 4--5; port customs holds 6--7, 12, 21, 26; supplier degraded wk~20 \\
112 & \textsc{isolated} & demand shifts only: decline 13--14, seasonal lift 20--24 \\
1   & \textsc{persistent} & port congested 17--24 $+$ demand seasonal lift 20--26 $+$ freight spike wk~19; demand promo 9--12 \\
172 & \textsc{persistent} & port congested 5--11 $+$ demand promo 9--12; demand seasonal lift 22--26 $+$ supplier degraded 24--26 \\
170 & \textsc{persistent} & freight spike 3--8 $+$ port congested 5--12; demand promos 2--5, 15--18; freight spike 24--26 $+$ supplier degraded wk~26 \\
21  & \textsc{compound} & demand seasonal lift 10--17; port customs hold wk~11; then canal closed 21--26 $+$ quality out of control 22--24 $+$ supplier degraded wk~23 \\
143 & \textsc{compound} & freight spike 5--10 running into port congestion 10--17; supplier degraded 2--5; port customs hold wk~1; canal closed wks 15, 25 \\
99  & \textsc{compound} & the marathon tape: two port-congestion runs plus a customs hold, two canal closures, three demand-regime changes, supplier degraded 23--24 \\
\bottomrule
\end{tabular}
\end{center}

\begin{figure}[t]
  \centering
  \includegraphics[width=\textwidth]{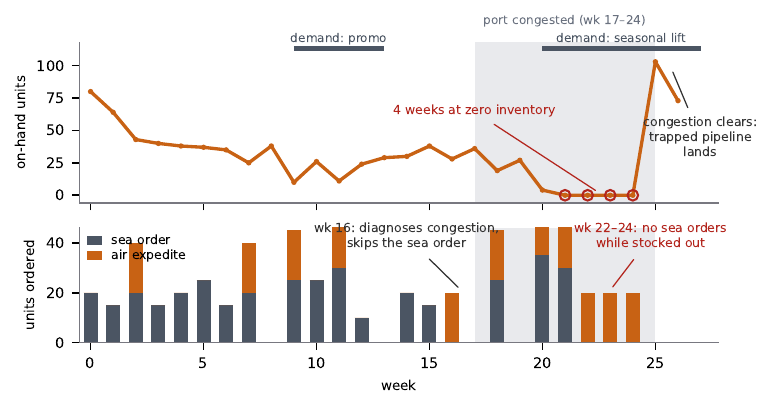}
  \caption{\textbf{The freeze, week by week (Claude Sonnet 5, seed 1).}
  Top: on-hand inventory; the shaded band is the true port-congestion window
  (weeks 17--24) and the rugs above mark the true demand windows, none of
  which the agent observes directly. Bottom: units ordered by sea and by
  air each week. The agent diagnoses the congestion in week 16, its sea
  orders pile up behind the blocked port, and in weeks 22--24 it stops
  ordering by sea while holding zero inventory --- reasoning that the
  trapped pipeline is ample cover. When the congestion clears in week 25,
  the trapped orders land at once. Every series is parsed from the raw
  episode trace.}
  \label{fig:seed1}
\end{figure}

\paragraph{Exhibit A: freezing (Claude Sonnet 5, seed 1).} Week 16, on
diagnosing the congestion: ``Port congestion has spiked severely
(berth\_wait 10 days\ldots) \ldots\ I'll skip placing a new sea order this
week \ldots\ and reassess next week once conditions hopefully ease.''
Week 22, with zero on-hand inventory and a stockout in progress: ``Despite
the current 0 on-hand inventory, I already have a very large pipeline: 104
units in transit (arriving weeks 23--25) \ldots\ roughly 4--5 weeks of cover
\ldots\ ordering more now would only add to an already oversized pipeline,
incurring extra holding/in-transit costs and demurrage risk given
berth\_wait is still elevated (15 days).'' Weeks 21--24 all close at zero
inventory (\$660--\$823 per week).

\paragraph{Exhibit B: flailing (DeepSeek-V4-Pro, seed 25).} Week 4:
``Week 4: Disruption confirmed --- Suez counts crashed to 28/25 \ldots\ Cape
is now the only safe route. Locked freight at 1.02x for 4 weeks. Ordering 40
via Cape/spot \ldots'' Week 6, correctly closing the diagnosis: ``Week 6: Suez
reopening --- counts back to 56/56 \ldots'' By then the freight lock, air
expedite, and stacked orders are sunk; weekly costs run \$610--\$814 through
week 8, and the episode ends at \$7{,}950 total cost, skill $-1.45$.

\section{Agent Interface: Prompt and Observation Format}
\label{app:prompt}

\paragraph{Weekly observation.} The observation dictionary below is
serialized into the agent's context each week, exactly as produced by the
environment (a presentation-manifest key used only by the web frontend is
omitted). This example is week 10 of seed 1 under the base-stock policy:
the demand promo is active (point-of-sale 29 against a baseline of 20), the
lane is calm, and the symptom-blind policy has just taken a small stockout.

{\footnotesize
\begin{verbatim}
{
 "week": 10,
 "inventory": 0,
 "on_order": 66,
 "inventory_position": 66,
 "arrived": 27,
 "components": {
  "unit": {
   "on_hand": 0,
   "on_order": 66,
   "inventory_position": 66,
   "arrived": 27
  }
 },
 "served": {
  "unit": 27
 },
 "pipeline": [
  {
   "qty": 23,
   "route": "suez",
   "supplier": "spot",
   "component": "unit",
   "dispatched_week": 8,
   "status": "at_sea",
   "eta": 11
  },
  {
   "qty": 15,
   "route": "suez",
   "supplier": "spot",
   "component": "unit",
   "dispatched_week": 9,
   "status": "at_sea",
   "eta": 12
  },
  {
   "qty": 28,
   "route": "suez",
   "supplier": "spot",
   "component": "unit",
   "dispatched_week": 10,
   "status": "at_sea",
   "eta": 13
  }
 ],
 "cost_breakdown": {
  "shipping": 80.192,
  "surcharge": 0.0,
  "holding": 0.0,
  "in_transit": 66.0,
  "stockout": 40.0,
  "couple": 0.0,
  "demurrage": 0.0,
  "rework": 0.0
 },
 "contracts": [
  {
   "supplier": "spot",
   "start_week": 0,
   "end_week": null,
   "unit_price": 4.0,
   "otif_floor": 85,
   "break_fee": 10.0
  }
 ],
 "contract_open": [],
 "term_menu": [
  "short",
  "long",
  "strict",
  "lenient"
 ],
 "suez_count": 70,
 "bab_count": 70,
 "cape_count": 60,
 "bulletin": "Shipping lanes normal. Suez Canal and Bab-el-Mandeb transits at
   seasonal levels; Cape routing remains a niche choice.",
 "suppliers": [
  {
   "id": "qualified",
   "otif": 99,
   "lead_days": 14,
   "band": "ontime",
   "onboard_lead": 0,
   "unit_premium": 1.0,
   "unit_delta": 1.0
  },
  {
   "id": "spot",
   "otif": 98,
   "lead_days": 14,
   "band": "ontime",
   "onboard_lead": 0,
   "realized_lead_slip": 0.0,
   "unit_discount": 1.5,
   "unit_delta": -1.5
  },
  {
   "id": "backup",
   "otif": 95,
   "lead_days": 16,
   "band": "ontime",
   "onboard_lead": 1,
   "unit_delta": 0.3
  }
 ],
 "pos_units": 29,
 "demand_forecast": 28,
 "freight_index": 109,
 "freight_outlook": 85,
 "berth_wait": 0,
 "wait_outlook": 8,
 "aql_result": "accept",
 "realized_fill": 1.0
}
\end{verbatim}}

\paragraph{System prompt (FAITHFUL arm).} The complete system prompt used
for all 200 runs is reproduced verbatim below. It specifies mechanics,
costs, tool semantics, and the latent structure of each factor, and
contains no strategy advice; its inclusion substantiates the rules-only
claim of Section~3.

{\footnotesize
\begin{verbatim}
You run the import replenishment desk for a European importer on the Asia-
Europe shipping lane. You will run it for 26 weeks. Your one objective is to
MINIMIZE TOTAL COST over the whole 26-week horizon -- not any single week.

THE WEEK
Each week you see the current situation, then make exactly one decision by
calling place_order. The world advances only when you call place_order. You
start with 80 units on hand. Demand is roughly 20 units a week (it drifts --
see DEMAND below), served from on-hand inventory; unmet demand is a stockout.

YOUR LEVERS (these are the only actions; mirror them exactly)
- place_order(rationale, qty, route, supplier, contract_action,
  contract_supplier, contract_terms): one weekly decision that may place an
  order, manage a contract, or both.
  - rationale (REQUIRED, every week): a few sentences working through THIS
week's decision and why this qty/route/supplier (and any contract). The world
does not advance without it.
  - qty: any whole number of units to order this week, from 0 up to 100. 0 =
order nothing (no route/supplier needed). There is no fixed menu.
  - route "suez" or "cape" (required if qty > 0):
    - "suez": base 4/unit, faster (~3 weeks), but the Suez/Red Sea corridor
can be disrupted -- a ship caught at the canal during a disruption waits, then
diverts around the Cape (arriving much later and billed the Cape price
difference).
    - "cape": base 6/unit, slower (~4 weeks), but it bypasses the Suez
corridor and is reliable.
  - supplier "qualified", "spot", or "backup" (required if qty > 0). You may
only source a supplier you hold a LIVE contract with -- see CONTRACTS and the
`contracts` / `contract_open` keys in the report.
  - to manage a contract THIS week, set contract_action ("sign", "switch",
"renew", or "lapse"), contract_supplier, and (for sign/switch/renew)
contract_terms ("short", "long", "strict", "lenient"). The contract resolves
BEFORE the order, so you can sign a supplier and source it in the same call.
Use qty 0 to manage a contract without ordering.
- buy_briefing(): pay 30 for a one-line analyst assessment of THIS week's LANE
  state (the Suez corridor), before you order. Optional.
- buy_audit(): pay 25 for a direct read of your spot supplier's current
  reliability state, before you order. Optional.
- lock_freight(weeks): forward-buy the freight rate -- FIX this week's freight
  cost multiplier for the next `weeks` weeks. While locked you pay the locked
  rate regardless of the spot index: it shields you from a rate spike, but you
  forgo a drop, and an unused week still burns the window. A within-week
  action (it does NOT advance the week); lock, then place_order in the same
  week to ship at the locked rate. The active lock shows as `freight_lock`
  (rate + weeks_left) in the report.
- expedite_air(qty): fly units in on a fast air lane that BYPASSES the
  destination port -- they land in your inventory NEXT week regardless of port
  congestion, at 15/unit, capped at 20 units a week. A within-week action (it
  does NOT advance the week); expedite, then place_order in the same week.
- inspect_batch(): pay 40 to run an incoming inspection on THIS week's
  arriving batch -- it sorts and reworks the defects, recovering about 70% of
  them before they stock (fewer units lost to defects, less rework). A within-
  week action (it does NOT advance the week); inspect, then place_order in the
  same week.

SUPPLIERS (who you buy from -- pick per order)
- qualified (the incumbent): reliable (99% OTIF), but dearest -- it adds
  1.0/unit over the route base (Suez 5/unit, Cape 7/unit). Always ships your
  full quantity. Evergreen contract. In this task you are NOT contracted to
  qualified at the start -- sign it to migrate off spot when you judge spot
  has turned.
- spot (YOUR STARTING INCUMBENT: you begin already contracted to it and source
  it by default): cheapest -- 1.5/unit BELOW the route base (Suez 2.5/unit,
  Cape 4.5/unit) -- but its reliability DRIFTS and you cannot see it directly.
  A healthy spot ships your full qty; a wobbling one ships only about HALF; a
  degraded one ships NOTHING; and it can go DEFUNCT (fail for good, gone for
  the rest of the horizon). Its OTIF scorecard (ontime / slipping / failing /
  defunct) is the contracted on-time metric; you also see your realized
  experience with spot -- realized_fill (how much of an order actually shipped
  when you sourced it) and realized_lead_slip (its reported lead-time this
  week), both noisy week to week. buy_audit gives a direct read of its current
  state. "slipping" is ambiguous -- it is either a wobble that recovers or the
  first week of a real failure; the following weeks tell you which. AND a spot
  shortfall DURING a lane disruption is back-ordered at a crisis rate about 3x
  a normal stockout.
- backup (a second qualified source): reliable (95% OTIF), a small premium
  (+0.3/unit over the route base), but it needs 1 week of onboarding before
  its FIRST order can ship.

CONTRACTS (the gate on sourcing)
- You can only source a supplier you currently hold a live contract with. The
  report shows your `contracts`, `contract_open` (contracts that have expired
  or whose supplier died -- these need renewing), and `term_menu`.
- contract_action "sign"/"switch"/"renew" opens a fresh contract on the chosen
  supplier with the chosen terms; "lapse" drops an open contract (surrender).
  A contract auto-opens when it expires or its supplier dies, and qualified's
  contract is evergreen.
- terms menu (the locked unit price is set off the Suez base, 4/unit):
    - "short": 4 weeks, ~3% cheaper, easy to exit.
    - "long": 12 weeks, ~6% dearer (a price-lock), hard to exit.
    - "strict": 8 weeks, high OTIF floor (the supplier owes you on a slip).
    - "lenient": 8 weeks, ~5% cheaper, no real penalty -- you eat the risk.
- Carrying 2 or more live contracts costs 4/week (dual-source overhead).

COSTS (every number is real; weigh them)
- shipping: the route base, adjusted for the supplier (above), then scaled by
  the freight index (see FREIGHT), paid when you order.
- holding: 1 per unit per week -- on inventory ON HAND and IN TRANSIT (capital
  on the water still costs you).
- stockout: 20 per unit of unmet demand in a week.
- surcharge: a Suez ship diverted around the Cape is billed the Cape-vs-Suez
  difference.
- demurrage: 2 per held unit per week when the destination port holds your
  arrivals (see PORT).
- air: 15 per unit when you expedite_air to fly units past a jammed port (see
  PORT).
- rework: 15 per defective unit when quality is off (see QUALITY).
- inspect: 40 when you inspect_batch to sort a bad arriving batch (see
  QUALITY).
- briefing: 30 each; dual-source overhead: 4/week.

WHAT YOU SEE EACH WEEK (your only signals; the latent ones are NOISY)
- week, inventory, arrived, and pipeline (your in-flight shipments with
  estimated arrival weeks).
- inventory_position and on_order: on_order is the total units already ordered
  but not yet arrived (your pipeline); inventory_position = inventory on hand
  + on_order.
- LANE: suez_count, bab_count, cape_count (ships that transited the Suez
  Canal, the Bab-el-Mandeb strait, and the Cape this week) plus a trade-press
  bulletin.
- SUPPLIERS: an OTIF scorecard per supplier (band + on-time % + quoted lead).
  For spot you also see realized_fill (actual vs ordered, when you sourced it)
  and realized_lead_slip (its reported lead behaviour this week).
- DEMAND: pos_units (what actually sold this week) and demand_forecast (a
  forward read). Both noisy. Underlying demand drifts between normal, short
  promo spikes, sustained seasonal lifts, and a sticky structural decline.
- FREIGHT: freight_index (this week's spot-rate level; ~100 is normal) and
  freight_outlook (a noisier forward read). Your shipping cost is scaled by
  the index the week you order. The underlying rate regime drifts between
  slack (cheap), normal, tightening, and a costly spike.
- PORT: berth_wait (days) and wait_outlook. When the destination port congests
  or a customs hold lands, your arrivals are held a week and accrue demurrage.
- QUALITY: aql_result (accept / marginal / reject -- incoming inspection).
  When the supplier's process drifts out of control, a fraction of your
  arrivals are defective: they do not stock and they cost rework.
- cost_breakdown: what last week cost you, by category.

THE LANE STRUCTURE (told to you plainly -- use it)
There is a disruption process on the Suez corridor that you cannot observe
directly. It builds up, may or may not break into a real disruption, and if it
does, the disruption is either short (clears within a couple of weeks) or long
(drags on for many weeks). You only learn which by tracking how the weekly
counts, the bulletin, and your shipment ETAs evolve over time.

When trouble first breaks, a genuine disruption and a false alarm look
IDENTICAL for one week -- the counts drop the same way whether it is nothing
or the first week of a real disruption. The ambiguity resolves the FOLLOWING
week: a false alarm snaps back to normal; a real disruption stays down, and
its counts then reveal whether it is the short or the long kind.

THE LOOP (you own it)
Run the full episode yourself. Each week: read the latest situation report (it
arrives with the kickoff and with every order you place), optionally use the
within-week levers, then call place_order exactly once -- with a written
`rationale` for that week's decision. Placing the order advances the world to
the next week and returns the new report. Keep going week after week until
place_order tells you the episode is done. Do NOT ask the human anything. Do
NOT stop early. Every week's reasoning goes in the place_order `rationale` so
your thinking is always visible.
\end{verbatim}}

\end{document}